\documentclass[hidelinks]{article} 
\pdfoutput=1
\usepackage{nips14submit_e,times}
\usepackage{url}
\usepackage[sort&compress]{natbib} 
\usepackage{graphicx}

\title{Unsupervised Feature Learning through Divergent Discriminative Feature Accumulation}

\author{
Paul A.~Szerlip, Gregory Morse, Justin K.~Pugh, and Kenneth O.~Stanley \\
Department EECS (Computer Science Divison)\\
University of Central Florida\\
Orlando, FL 32816 USA\\
\texttt{\{pszerlip,jpugh,kstanley\}@eecs.ucf.edu, gregorymorse07@gmail.com}
}

\nipsfinalcopy 

\begin{document}

\maketitle

\begin{abstract}
Unlike unsupervised approaches such as autoencoders
that learn to reconstruct their inputs,
this paper introduces an alternative approach to 
unsupervised feature learning called \emph{divergent discriminative
feature accumulation} (DDFA)
that instead continually accumulates features that
make novel discriminations among the training set.  Thus DDFA features
are inherently discriminative from the start even though they are
trained without knowledge of the ultimate classification problem.
Interestingly, DDFA also continues to add new features indefinitely
(so it does not depend on a hidden layer size),
is not based on minimizing error, and is inherently divergent instead
of convergent, thereby providing a unique direction of research
for unsupervised feature learning.  In this paper the quality of
its learned features is demonstrated on the MNIST dataset,
where its performance confirms that indeed DDFA 
is a viable technique for learning 
useful features.                     
\end{abstract}

\section{Introduction}



The increasing realization in recent years that artificial neural
networks (ANNs) can learn many layers of features
\citep{bengio:nips07,hinton:nc06,marcaurelio:nips07,ciresan:nc10} 
has reinvigorated the study of representation learning in ANNs 
\citep{bengio:ieeetpam13}.  While the beginning of this
renaissance focused on the sequential unsupervised
training of individual layers one upon 
another \citep{bengio:nips07,hinton:nc06}, the number of approaches
and variations that have proven effective at training in such
\emph{deep learning} has since 
exploded \citep{schmidhuber:agi11,bengio:ieeetpam13}.  
This explosion has in
turn raised the question of what makes a good representation,
and how it is best learned \citep{bengio:ieeetpam13}.
The main contribution of this paper is to advance our understanding
of good representation learning by suggesting a new principle
for obtaining useful representations that is accompanied by
a practical algorithm embodying the principle. 

The feature representation obtained through learning
algorithms is often impacted by the nature of the training.
For example, supervised approaches 
such as stochastic gradient descent \citep{ciresan:nc10}
that aim to minimize the error in a classification problem 
in effect encourage the exclusive discovery of features that 
help to discriminate
among the target classifications.  
In contrast, unsupervised approaches, which include
both generative representations such as 
restricted Boltzmann machines (RBMs) \citep{hinton:nc06}
and autoencoders that are trained to reproduce
their inputs \citep{bengio:nips07}, yield a more 
general feature set that captures dimensions of variation
that may or may not be essential to the classification objective.
The hope of course is that such a set would nevertheless be
useful for classification in any case, and the pros and cons 
of e.g.\ generative versus discriminative features have
proven both subtle and complex \citep{ng:nips02,jaakkola:nips98}. 
Nevertheless, one benefit of unsupervised training is that it does
not require labeled data to gain traction.  

An important insight in this paper is that there is an 
unrecognized option
outside this usual unsupervised versus supervised (or generative versus
discriminative) dichotomy. 
In particular, there is an alternative kind of discriminative learning
that is unsupervised rather than supervised. 
In this proposed alternative 
approach, 
called  \emph{divergent discriminative
feature accumulation} (DDFA),
instead of searching for features constrained by
the objective of solving the discriminative classification problem,
a learning algorithm can instead attempt to \emph{collect} as many
features that discriminate strongly among training examples 
as possible, without regard to any particular classification problem.

The approach in such unsupervised discriminative learning is thus to search 
continually for novel features that discriminate among
training examples in new ways.  Interestingly,
unlike conventional algorithms in deep
learning, such a search is explicitly 
divergent by design and therefore continues to
accumulate new features without converging.   
In effect, a high-scoring feature is therefore
relevant to discriminating among the examples,
even though the ultimate discrimination problem is not known.
A comprehensive set of such 
features that discriminate
among the training set in fundamental ways is thereby suitable 
in principle for
later supervised training \emph{from those collected features} for
any particular discrimination task.
This idea is intuitive in the sense that even for humans,
distinctions among experiences
can be learned before we know how we will apply such distinctions.
Furthermore, as the results will show, if the search gradually shifts
from simple to more complex distinctions, only a small subset
of all possible distinctions (many of which are obscure) needs to be 
discovered.


In fact,                
this perspective on feature learning has the advantage 
over more conventional approaches in deep learning that
learning does not depend on a fixed a priori number
of features.
Rather, it simply continues to accumulate new features 
as long as the algorithm is run.  
Furthermore, unlike in other unsupervised approaches, the
accumulated features are known explicitly to be discriminative,
suiting them well to later discriminative learning.   
Another potentially advantageous property of such a feature accumulator
is its lack of convergence (thereby avoiding the problem of local optima), 
which stems from the fact that it is inherently divergent  
because it is not based on 
minimizing an error.
In these ways DDFA is
uniquely flexible and autonomous.   

The driving force behind the feature accumulator is the 
imperative of finding novel features.
Thus a well-suited algorithm for implementing this idea in
practice is the recent \emph{novelty search} algorithm
\citep{lehman:ecj11}, which is 
a divergent evolutionary algorithm that is rewarded
for moving \emph{away} in the search space of candidate behaviors 
(such as discriminations)
from where it has already visited to where it has not. 
By accumulating features that are themselves
ANNs, novelty search in this paper enables 
DDFA.
As with other unsupervised pretraining approaches such as autoencoders,
once DDFA determines that sufficient features are collected,
a classifier is trained above them for the classification
task (through backpropagation in this paper). 
To demonstrate the potential of DDFA to collect useful features,
it is tested in this paper by collecting single-layer
features for the MNIST digital handwriting recognition 
benchmark \citep{lecun:mnist98}.
Even with the consequent two-layer shallow classifier network, its   
testing performance rivals more conventional training techniques. 

This initial proof of concept establishes the efficacy of
accumulating features as a basis for representation learning.
While the simple one-layer accumulated discriminative features
from DDFA perform well,
DDFA can conceivably improve further through layering
(e.g.\ accumulating multilayer features or
searching for novel 
features that are built above already-discovered
features) and convolution \citep{lecun:hbtnn95}, 
just like other deep learning
algorithms.  
Most importantly,
based on the novel representational principle 
that discriminators make 
good features
for classification problems, DDFA opens up
a new class of learning approaches.

\section{Background}

This section reviews the two algorithms, novelty search
and HyperNEAT, that underpin the DDFA approach. 

\subsection{Novelty Search} 

The problem of \emph{collecting} novel instances of a class 
is different from the 
more familiar problem of minimizing error.  
While error minimization aims at converging towards 
minima in the search space, collecting novelty requires
diverging away from past discoveries and fanning out
across the search space in all directions that appear to lead 
towards further novelty.  
This fanning-out process is thus
well-suited to a population-based approach that accumulates
and remembers novel discoveries to help 
push the search continually to even more novelty as it progresses.
The novelty search algorithm \citep{lehman:ecj11} 
implements such a process in practice through an evolutionary
approach, which naturally provides the population-driven context
appropriate for finding novelty.  However, it is important to note
that novelty search is unlike even traditional evolutionary 
algorithms (EAs), which themselves are usually driven to
converge to higher fitness. 
In fact, while EAs are often viewed as 
an alternative approach to optimization, their natural capacity
to \emph{diversify and collect} may instead better capture their
practical potential to contribute to problems in learning. 

The idea in novelty search is to reward candidates (by increasing
their probability of reproduction) who are behaviorally novel.  If the
candidates are ANNs as in the present study, then the word
``behaviorally'' becomes critical because it refers to what the
discovered ANNs \emph{actually do} (e.g.\ how they discriminate)
as opposed to just their underlying
genetic representations (i.e.\ genomes), which may or may not do
anything interesting.  Thus discovering novel behaviors 
requires search (as opposed to just enumerating random sets of genes),
thereby instantiating a nontrivial 
alternative to the traditional objective gradient.

This point is particularly important in the context of deep learning,
where researchers have commented on the potential 
long-term limitations of optimization gradients 
and the need for a broader and less convergent approaches for learning
representations.  For example, when discussing the
future of representation learning, \citet{bengio:slsp13} notes:

\begin{quotation}
The basic idea is that humans (and current learning algorithms) are
limited to ``local descent'' optimization methods, that make small
changes in the parameter values with the effect of reducing the
expected loss in average. This is clearly prone to the presence of
local minima, while a more global search (in the spirit of both
genetic and cultural evolution) could potentially reduce this
difficulty. 
\end{quotation}
   
Novelty search \citep{lehman:ecj11} can be viewed as an embodiment of just such 
a ``genetic evolution'' that is suited to accumulating discoveries
free from the pitfalls of ``local descent.''  In fact,
while novelty search was originally shown sometimes to 
find the objective of an optimization
problem more effectively than objective-based optimization 
\citep{lehman:ecj11}, 
\citet{cully:gecco13} recently raised the intriguing 
notion of novelty search as a \emph{repertoire collector}.
That is, instead of searching for a \emph{solution} to a problem,
novelty search can collect a set of novel skills (each one a point in 
the search space) intended for later aggregation by a higher-level mechanism.
This repertoire-collecting idea aligns elegantly with 
the problem of accumulating features for deep learning, wherein each
feature detector can be viewed as a ``skill'' within the repertoire of
a classifier.

In practice, novelty search maintains an \emph{archive} of previously
novel discoveries as part of the algorithm.  Future candidates are
then compared to the archive to determine whether they too are novel.
A random sampling of candidates is entered into the archive,
which implies that more frequently-visited areas will be
more densely covered. 
Intuitively, if the
average distance to the nearest neighbors of a given behavior $b$ 
is large then it
is in a sparse area; it is in a dense region if the average distance
is small. The sparseness ${\rho}$ at point ${b}$ is given by
\begin{eqnarray} \label{eq1}
\rho(x) = \frac{1}{k}\sum_{i=0}^k \mathrm{dist}(b,{\mu_{i}}),
\end{eqnarray}
where ${\mu_{i}}$ is the ${i}$th-nearest neighbor of ${b}$ with
respect to the distance metric \emph{dist}, which is a
domain-dependent measure of behavioral difference between two
individuals in the search space.
The nearest neighbors calculation
must take into consideration individuals from the current population
and from the permanent archive of novel individuals.
Candidates from
more sparse regions of this behavioral search space then receive
higher novelty scores, which lead to a higher chance of reproduction.  
It is important to note that this novelty space
cannot be explored purposefully, that is, it is not known \emph{a
  priori} how to enter areas of low density just as it is not known a
priori how to construct a solution close to the objective. Thus,
moving through the space of novel behaviors requires exploration. 
The gradient of novelty is interesting in particular because novel
discoveries lead to other novel discoveries, which means that a search algorithm
following gradients of novelty is likely to make many interesting discoveries. 

Novelty search in effect runs as a regular EA wherein novelty replaces fitness
as the criterion for selection, and an expanding archive of past novel discoveries              
is maintained.  This simple idea 
will empower DDFA in this paper to accumulate a collection of novel features.

\subsection{HyperNEAT}  

The term for algorithms that search for ANNs through an
evolutionary process is \emph{neuroevolution}
\citep{stanley:ec02,floreano:ei08}.  It is important to note
that modern neuroevolution algorithms are not like conventional EAs
based on bit strings, but instead implement a variety of sophisticated
heuristics and encodings that enable the discovery of large and
well-organized networks.  This section is designed to introduce the
particular neuroevolution algorithm (called HyperNEAT) that is
combined with novelty search to search for features in this
paper.  Because neuroevolution is an independent field 
that may be unfamiliar to many in deep learning,
this section is written to emphasize the main ideas that make it
appealing for the purpose of feature learning, without including details
that are unnecessary to understand the operation of the proposed
DDFA algorithm.  
The complete details of HyperNEAT can be found in its primary 
sources \citep{stanley:alife09,gauci:nc10,gauci:aaai08,verbancsics:jmlr10}.
  
\begin{figure}
\begin{center}
  \begin{minipage}{1.6in}  
     \centering
     \includegraphics[height=0.4in]{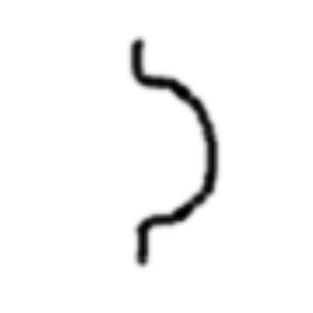}

     (a) Original Weight Pattern
  \end{minipage}
  \begin{minipage}{1.6in}  
     \centering
     \includegraphics[height=0.4in]{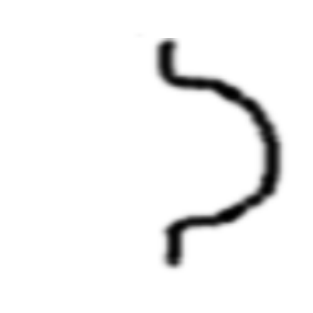}

     (b) HyperNEAT Mutation 
  \end{minipage}
  \begin{minipage}{1.6in}  
     \centering
     \includegraphics[height=0.4in]{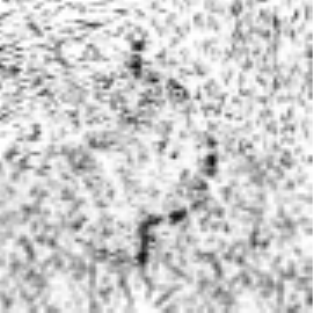}

     (c) Uniform Mutation
  \end{minipage}
\end{center}
\caption{\textbf{Systematic Deformation in HyperNEAT-style Mutation.}
Each image depicts the pattern of weights projecting from a
single $28\times 28$ input field to a single output node.
The weight of a hypothetical feature (a) exhibits contiguity and 
some symmetry.  The HyperNEAT style of mutation (b) perturbs
the pattern of weights while still preserving the geometric
regularities of the original feature.  In contrast, simply
randomly mutating weights with uniform probability (c) leads
to an incoherent corruption of the original feature.   
}
\label{fg:deformation}
\end{figure}

In a domain like visual recognition, 
the pattern of weights in useful features 
can be expected to exhibit a degree of contiguity and perhaps regularity.
For example, it is unlikely that an entirely random pattern of largely 
unconnected pixels corresponds to a useful or interesting feature.  
It has accordingly long been
recognized in neuroevolution that entirely random perturbations of
weight patterns, which are likely to emerge for example from random
mutations, are unlikely to maintain contiguity or regularity.  While
stochastic gradient descent (SGD) algorithms at least justify their 
trajectory through the search space through descent, a completely
random perturbation of weights is arguably less principled and therefore
perhaps less effective.  Nevertheless, SGD still suffers to some extent
from the same problem that even a step that reduces error may not
maintain contiguity or regularity in the feature geometry.  
Neuroevolution algorithms have responded to this concern with
a class of representations called 
\emph{indirect encodings} \citep{stanley:alife03},
wherein the weight pattern is generated by an evolving genetic encoding
that is biased towards contiguity and regularity by design.             
That way, when a mutation is applied to a feature, the feature 
deforms in a systematic though still randomized fashion 
(figure \ref{fg:deformation}).

HyperNEAT, which stands for \emph{Hypercube-based NeuroEvolution of
Augmenting 
Topologies} \citep{stanley:alife09,gauci:nc10,gauci:aaai08,verbancsics:jmlr10}
is a contemporary neuroevolution algorithm
based on such an indirect encoding.  In short, HyperNEAT evolves an
encoding network called a compositional pattern producing network
(CPPN; \citealt{stanley:gpem07}) 
that \emph{describes} the pattern of connectivity within
the ANN it encodes.
Therefore, mutations in HyperNEAT
happen to the \emph{CPPN}, 
which then transfers their effects to the
encoded ANN.  In this way the CPPN is like DNA, which transfers
the effects of its own mutations to the structures it encodes,
such as the brain.  
Because the CPPN encoding is designed to describe patterns of
weights across the geometry of the encoded network,
the weights in HyperNEAT ANNs
tend to deform in contiguity-preserving and regularity-preserving ways
(as seen in figure \ref{fg:deformation}),
thereby providing a useful bias \citep{stanley:alife09,gauci:nc10}. 
Furthermore, CPPNs in HyperNEAT grow over evolution
(i.e.\ their structure is augmented over learning), which means that the 
pattern of weights in the ANN they describe (which is fixed in size)
can become more intricate and complex over time.
For unfamiliar readers, 
it is worth noting that this
HyperNEAT style of representation for ANNs
is well-established and has appeared in mainstream venues such as 
AAAI \citep{gauci:aaai08}, Neural Computation \citep{gauci:nc10},
and JMLR \citep{verbancsics:jmlr10}. 

An important observation is that HyperNEAT's tendency to preserve
geometric properties in its weights means that it is
not invariant to permutations in the input vector.  In effect 
(in e.g.\ the case of MNIST) it
is exploiting the known two-dimensional geometry of the problem.
However, at the same time, while it does exploit geometry, 
its use in this paper is
not convolutional either: its input field is never
broken into receptive fields and is 
rather projected in whole directly (without intervening layers)    
to a single-feature output node.  Thus the powerful advantage 
of convolution for visual problems is \emph{not available in
this investigation}, making the problem more challenging.
As a consequence, the DDFA implementation in this paper does not
fit neatly into the permutation-invariant-or-not dichotomy,
and may be considered somewhere closer to typical
permutation-invariant scenarios.        

This overview of HyperNEAT is left brief because 
its other details (which are widely disseminated in the venues above)
are not essential to the main idea in this paper, which is
to accumulate feature detectors through novelty search.  

\vspace{-0.1in}
\section{Approach: Divergent Discriminative Feature Accumulation (DDFA)}

Unsupervised pretraining in deep learning has historically focused
on approaches such as autoencoders and RBMs \citep{bengio:nips07,hinton:nc06}
that attempt to \emph{reconstruct} training examples
by first translating them into a basis of features different
from the inputs, and then from those features regenerating the inputs.
This idea is appealing because the imperative of reconstruction
demands that the learned features must ultimately reflect some aspect of the
underlying structure of the training set.
Of course, one potential problem with this approach is that there
is no assurance that the learned features actually align with
any particular classification or discrimination problem for which they  
might be used in the future.    
Yet this conventional approach to learning features also raises some
interesting deeper questions.  For example, is there any \emph{other}
way to extract meaningful features and thereby learn representations
from a set of examples without explicit supervision?   
    
There are some well-known simple alternatives, though they are
not usually characterized as feature-learning algorithms.
For example clustering algorithms such as K-means or Gaussian
mixture models in effect extract structure from data that can
then assist in classification; in fact at least one study has
shown that such clustering algorithms can yield features as
effective or more so for classification than autoencoders or RBMs
\citep{coates:icais11}.  This result highlights that 
reconstruction is not the only effective incentive for extracting
useful structure from the world.  

The approach introduced here goes beyond simple clustering by
emphasizing the general ability to learn diverse \emph{distinctions}.
That is, while one can learn how to \emph{describe} the world,
one can also learn how different aspects of the world
\emph{relate} to each other.  Importantly, there is no
single ``correct'' view of such relations.  Rather,
a rich set of learned relationships can support drawing 
important distinctions later.  For example, in one view
palm trees and ``regular'' trees share properties that distinguish them
from other plants.  However, in another view, palm trees
are in fact distinct from regular trees.  \emph{Both} such views can be useful
in understanding nature, and one can hold both simultaneously
with no contradiction.  When an appropriate question comes
up, such as which plants are tall and decorative, the feature
\emph{tall} becomes available because it was learned to
help make such general distinctions about the world 
in the past.        
  
The idea in DDFA is to continually accumulate such distinctions
systematically through novelty search, thereby building an
increasingly rich repertoire of features that help divide
and relate observations of the world.  
Specifically, suppose there are $n$ training examples
$\{x^{(1)},...,x^{(n)}\}$;
whether or not they are labeled will not matter because  
feature learning will be unsupervised.
Suppose also that any single \emph{feature detector}
$h_i$ (i.e.\ a single hidden node that detects a particular feature)
outputs a real number whose intensity represents the degree to which
that feature is present in the input.  
It follows that $h_i$ will assign a real number $h_i^{(t)}$ to
every example $x^{(t)}$ depending on the degree to which $x^{(t)}$ contains
the feature of interest for $h_i$.
The output of $h_i$ for all features $x^{(t)}$ 
where $t = 1,\dots,n$
thereby
forms a vector $\{h_i^{(1)},\dots,h_i^{(n)}\}$ that can be 
interpreted as the \emph{signature} of feature detector $h_i$
across the entire training set.  
In effect the aim is 
to continually discover new such signatures.

This problem of continually discovering novel signatures is naturally
captured by novelty search, which can be set up explicitly to
evolve feature detectors $h_i$, each of which takes a training
example as input and returns a single output.  The signature
$\{h_i^{(1)},...,h_i^{(n)}\}$ of $h_i$ over all training examples
is then its \emph{behavior characterization} for novelty search.
The novelty of the signature is then measured by comparing it
to the $k$-nearest signatures in the novelty archive, following
equation \ref{eq1}.  Novelty search then dictates that more novel features
are more likely to reproduce, which means that gradients of
novel signatures will be followed in parallel by the evolving
population.  Those features whose sparseness ${\rho}$
(i.e.\ novelty) exceeds a minimum threshold 
${\rho}_{\textrm{\footnotesize min}}$ 
are stored in the growing novel feature collection for later 
classifier training.

A likely source of confusion is the question of whether DDFA is a kind
of exhaustive search over signatures, which would not tend to 
discover \emph{useful} features in a reasonable runtime. 
After all, the number of theoretically possible distinctions is
exponential in the number of training examples.   
However, a critical facet of novelty-based searches that are combined
with HyperNEAT-based neuroevolution
is that the complexity of features (and hence distinctions) 
tends to increase 
over the 
run \citep{lehman:ecj11}.
As a result, the initial features discovered encompass
simple principles (e.g.\ is the left side of the image dark?) that
gradually increase in complexity.  For this reason, the most
arbitrary and incoherent features (e.g.\ are there 17 particular dots at 
specific non-contiguous coordinates in the image?) are possible
to discover only late in the search.  
Furthermore, because the novelty signature is measured over the training set,
features that make broad separations 
\emph{relevant}
to the training set itself  
are more likely to be discovered early.  In effect, over the course of DDFA,
the feature discoveries increasingly shift from simple
principles to intricate minutia.  Somewhere along this road are likely 
diminishing returns, well before all possible signatures are discovered.     
Empirical results reported here support this point.

Interestingly, because DDFA does not
depend on the minimization of error, in principle it can continue
to collect features virtually indefinitely, but in practice
at some point its features are fed into a classifier that is
trained from the collected discriminative features.

\vspace{-0.1in}
\section{Experiment}
\vspace{-0.1in} 

The key question addressed in this paper
is whether a divergent discriminative feature accumulator 
can learn \emph{useful} features, which means they should
aid in effective generalization on the test set.
If that is possible, the implication is that DDFA is a viable
alternative to other kinds of unsupervised pretraining.
To investigate this question DDFA is trained and tested on the MNIST
handwritten digit recognition dataset \citep{lecun:mnist98}, 
which consists of 60,000 training images and 10,000 test images.
Therefore, the signature of each candidate feature discovered by
DDFA during training is a vector of 60,000 real values.    

Because the structure of the networks that
are produced by HyperNEAT can include as many hidden layers as
the user chooses,
the question arises how many hidden layers should be allowed
in \emph{individual features} $h_i$ learned by HyperNEAT.
This consideration is substantive because in principle
DDFA can learn arbitrarily-deep individual features all at once,
which is unlike e.g.\ the layer-by-layer 
training of a deep stack of autoencoders.  
However, the explicit choice was made in this introductory experiment to
limit DDFA to single-layer features (i.e.\ without hidden nodes)
to disentangle the key question of whether the DDFA process represents
a useful principle from other questions of representation such
as the implications of greater depth.  Therefore,
feature quality is addressed straightforwardly in this study by
observing the quality of classifier produced based only on 
single-layer DDFA features.  As a result, 
the final classifier ANN has just two layers: the layer
of collected features and the ten-unit output layer for classifying
MNIST digits.       
    
The single-layer DDFA approach with novelty search and HyperNEAT is 
difficult to align directly with common deep learning approaches
in part because of its lack of permutation invariance even though
it is not convolutional in any sense (thereby lacking the
representational power of such networks), 
and its lack of depth in this initial test.  
Thus to get a fair sense of whether DDFA
learns useful features it is most illuminating to contrast it with
the leading result on an equivalently shallow two-layer architecture 
(which are rare in recent years) 
that similarly avoided special preprocessing like elastic distortions 
or deskewing.
In particular,  \citet{simard:icdar03} obtained one of the best
such results of 1.6\% test error performance.  
Thus a significant improvement on that result would suggest
that DDFA generates useful features that help to stretch the capacity
of such a shallow network to generalize. 
DDFA's further ability to approach the performance of conventional vanilla deep
networks, such as the original 1.2\% result from \citet{hinton:nc06}
on a four-layer network pretrained by a RBM,
would hint at DDFA's potential utility in the future for pretraining 
deeper networks. 

    
During the course of evolution, features are selected for reproduction
based on their signature's novelty score (sparseness $\rho$)
calculated as the sum of the distances to the k-nearest neighbors
($k=20$), where neighbors include other members of the population as
well as the historical novelty archive. At the end of each generation,
each individual in the population ($\textrm{size}=100$) has a $1\%$ chance of
being added to the novelty archive, resulting in an average of 1
individual added to the novelty archive on each generation.
Separately, a list of individuals called the \emph{feature list} is
maintained.  At the end of each generation, each member of the
population is scored against the current feature list by finding the
distance to the nearest neighbor ($k=1$), where neighbors are members
of the feature list. Those individuals that score above a threshold
${\rho}_{\textrm{\footnotesize min}} = 2{,}000$ are added to the feature list.  In effect, the
feature list is constructed in such a way that all collected features
have signatures that differ by at least ${\rho}_t$ from all others in
the collection. This threshold-based collection process protects
against collecting redundant features.
A simple variant of HyperNEAT called HyperNEAT-LEO \citep{verbancsics:gecco11}
(which leads to less connectivity) was the main neuroevolution engine.  
The HyperNEAT setup and 
parameters can be easily reproduced in full because they are simply
the default parameters of the SharpNEAT 2.0 publicly-available package
\citep{green:SharpNEAT}.

To observe the effect of collecting different numbers of features,  
DDFA was run separately until both 1,500 and 3,000
features were collected.  After collection concludes,
a set of ten classification nodes
is added on top of the collected features, and
simple backpropagation training commences.
The training and validation procedure mirrors that followed by
\citet{hinton:nc06}: first training is run on 50,000 examples
for 50 epochs
to find the network that performs best on a 10,000-example
validation set.  Then training shifts to the full 60,000-example
set, which is trained until it reaches the same training error as
in the best validation epoch.  The resulting network is finally tested
on the full 10,000-example test set.        
This whole procedure is similar to how autoencoders are trained before
gradient descent in deep learning \citep{bengio:nips07}.

\section{Results}


The main results are shown in Table \ref{fig:testingResults}.
DDFA was able to achieve test errors of 1.42\% and 1.25\% 
from collections of 1,500 and 3,000 features, respectively,
which are both well below the 1.6\% error of the similar 
shallow network trained without preprocessing from \citet{simard:icdar03}.
In fact, the result for the 3,000-feature network even approaches
the 1.2\% error of the significantly deeper network of \citet{hinton:nc06},
showing that shallow networks can generalize surprisingly well
by finding sufficiently high-quality feature sets, even despite 
a lack of exposure to distortions during training.    
It also appears that more collected features lead to better generalization, at
least at these sizes.  
It took 338 and 676 generations of feature collection
to obtain the 1,500 and 3,000 features, respectively.  Collecting 3,000 
features took about 36 hours of computation on 12 3.0 GHz cores.  

\begin{table}
\centering
    \begin{tabular}{| l | c | c | c |}
    \hline   
    \textbf{Features} & \textbf{DDFA Test Error} & \textbf{Random CPPNs Control} & \textbf{Random Weights Control} \\ \hline 
    \textbf{1,500} & 1.42\% & 1.63\% & 2.21\% \\ \hline
    \textbf{3,000} & \textbf{1.25\%} & 1.61\% & 2.00\% \\ \hline
    \end{tabular}
\caption{\textbf{MNIST Testing Error Rates of DDFA and Controls.}
}
\label{fig:testingResults}
\end{table}

\begin{figure}
\begin{center}
     \includegraphics[height=0.4in]{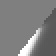}
     \includegraphics[height=0.4in]{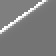}
     \includegraphics[height=0.4in]{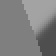}
     \includegraphics[height=0.4in]{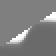}
     \includegraphics[height=0.4in]{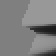}
     \includegraphics[height=0.4in]{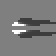}
     \includegraphics[height=0.4in]{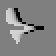}
     \includegraphics[height=0.4in]{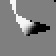}
     \includegraphics[height=0.4in]{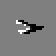}
     \includegraphics[height=0.4in]{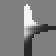}
\end{center}
\caption{\textbf{Example Collected Features.}
Each square is a weight pattern for an actual feature discovered by DDFA 
in which white is negative and black is positive.    
As these examples show, features range from simple line orientation detectors
reminiscent of those found in the lower levels of conventional deep
networks (towards left) to more complex shapes (towards right).        
}
\label{fg:features}
\end{figure}

Figure \ref{fg:features} shows a typical set of features 
collected by DDFA.  Interestingly, unlike the bottom layer of
deep learning networks that typically exhibit various line-orientation
detectors, DDFA also collects more complex features because newer features
of increasing complexity evolve from older features.   

To rule out the possibility that the reason for the testing performance 
is simply the HyperNEAT-based encoding of features,    
a \textbf{random CPPN control} was also run.  It follows an identical
procedure for training and testing, except that novelty scores
and adding to the feature list
during the feature accumulation phase are decided 
\emph{randomly}, which means the final collection in effect
contained random features with a range of CPPN complexity similar to the 
normal run.  To further investigate the value of the HyperNEAT
representation, an additional 
\textbf{random weights control} was tested
whose weights were assigned from a uniform random distribution, 
bypassing
HyperNEAT entirely.  
As the results in Table \ref{fig:testingResults} show,
the CPPN encoding in HyperNEAT provides a surprisingly good basis
for training even when the features are entirely randomly-generated.
However, they are still inferior to the features collected by 
normal DDFA.  As shown in the last column, without HyperNEAT, testing
performance with a collection of random features is unsurprisingly poor. 
In sum these controls show that the pretraining in DDFA is essential
to priming the later classifier for the best possible performance.        
      

\section{Discussion and Future Work}

The results suggest that DDFA can indeed collect useful features
and thereby serve as an alternative unsupervised feature learner.
While it may ultimately
lead to better training performance in some 
cutting-edge problems, 
future work with more layers and on larger problems 
is clearly necessary to investigate its full potential for
exceeding top results.

However, it is important to recognize that significantly more
than performance is at stake in the dissemination of
alternative unsupervised training techniques based on
new principles.  Deep learning faces several fundamental
challenges that are not only about testing performance.       
For example, recent surprising 
results from \citet{szegedy:corr13} 
show that very small yet anomalous perturbations of 
training images that are imperceptible to the human eye
can fool several different kinds of deep networks 
that nevertheless ominously score well on the 
test set.  The implications of these anomalies are
not yet understood. 
At the same time, as \citet{bengio:slsp13}
points out, local descent on its own will not 
ultimately be enough to tackle the most challenging problems,
suggesting the need for radical new kinds of optimization
that are more global.  These kinds of 
considerations suggest that simply scoring well on
a test set in the short run may not necessarily foreshadow
continuing success for the field in the long run.      

Therefore, the expanded possibilities that a validated
new principle  
can inspire are essential to the health
of an evolving field, whether or not it ultimately
breaks a particular benchmark record.
For example, DDFA shows that unsupervised
discriminative learning is possible and can be
effective, bringing with it several
intriguing corrollaries.  Among those,
it is possible to conceive training methods that
act as continual feature accumulators that do not
require a fixed ``hidden layer size.''   
Furthermore, it is possible to learn useful features
without any kind of error minimization
(which is even used in conventional unsupervised 
techniques).  Relatedly, an interesting question is whether
anomalous results are sometimes a side effect of the
very idea that all useful knowledge ultimately 
must come from minimizing error.   
The divergent dynamics of novelty search also
mean that the search is inherently more global
than local descent for the very reason that  
it is continually 
diverging, thereby offering a hint
of how more expansive feature sets can be
collected.
Thus, in addition to the many possibilities
for training multilayer deep features in DDFA,
another important path for future work is 
to investigate the long-term implications of 
these more
subtle differences from conventional techniques,
and to determine whether similar such unique 
properties can be introduced to deep learning through 
non-evolutionary techniques that follow gradients other than error. 
   

\bibliographystyle{nnapalike}
\bibliography{nnstrings,nn,newentries,mybib,ucf}

\appendix



%
%
%
%

\end{document}